\documentclass[10pt,conference]{IEEEtran}
\IEEEoverridecommandlockouts
\usepackage{cite}
\usepackage{amsmath,amssymb,amsfonts}
\usepackage{algorithmic}
\usepackage{graphicx}
\usepackage{textcomp}
\usepackage{subcaption}
\usepackage{xcolor}
\usepackage{float}
\usepackage{multirow}
\usepackage{hyperref}
\def\BibTeX{{\rm B\kern-.05em{\sc i\kern-.025em b}\kern-.08em
    T\kern-.1667em\lower.7ex\hbox{E}\kern-.125emX}}

\title{Spiking and Event-driven Neuromorphic Mamba Models for Efficient Speech Recognition}

\author{
\IEEEauthorblockN{
Tauseef Ahmed\IEEEauthorrefmark{1}\IEEEauthorrefmark{2}\IEEEauthorrefmark{3},
Tao Sun\IEEEauthorrefmark{1},
Jeronimo~Castrillon\IEEEauthorrefmark{3}\IEEEauthorrefmark{4},
Kanishkan Vadivel\IEEEauthorrefmark{2},
Guangzhi Tang\IEEEauthorrefmark{1}
}
\IEEEauthorblockA{\IEEEauthorrefmark{1}
Department of Advanced Computing Sciences, Maastricht University, Netherlands}
\IEEEauthorblockA{\IEEEauthorrefmark{2}
Hardware-Efficient AI Team, IMEC, Netherlands}
\IEEEauthorblockA{\IEEEauthorrefmark{4}
Center for Scalable Data Analytics and Artificial Intelligence (ScaDS.AI), Dresden, Germany}
\IEEEauthorblockA{\IEEEauthorrefmark{3}
Chair for Compiler Construction, TU Dresden, Germany}
}

\usepackage{tikz}
\usepackage{textcomp}
\usepackage{hyperref}
\usepackage{lipsum}

\newcommand\copyrighttext{%
  \footnotesize \textcopyright 2026 IEEE. Personal use of this material is permitted.
  Permission from IEEE must be obtained for all other uses, in any current or future
  media, including reprinting/republishing this material for advertising or promotional
  purposes, creating new collective works, for resale or redistribution to servers or
  lists, or reuse of any copyrighted component of this work in other works.}
\newcommand\copyrightnotice{%
\begin{tikzpicture}[remember picture,overlay]
\node[anchor=south,yshift=10pt] at (current page.south) {\fbox{\parbox{\dimexpr\textwidth-\fboxsep-\fboxrule\relax}{\copyrighttext}}};
\end{tikzpicture}%
}

\begin{document}

\maketitle
\copyrightnotice

\begin{abstract}
Deep learning has greatly advanced automatic speech recognition (ASR), enabling widespread deployment on edge devices such as smartphones and smart home systems. However, the computational and energy demands of deep neural networks pose significant challenges for such resource-constrained deployments, introducing latency and limiting real-time interaction. Neuromorphic computing offers a promising solution by introducing activation sparsity through spiking neural networks (SNNs) and event-driven neural networks, converting dense operations into sparse computations. However, a study that evaluates the hardware benefits of different neuromorphic strategies remains lacking for ASR. This paper explores spiking and event-driven neuromorphic neural networks to improve activation sparsity in the state-of-the-art SpeechMamba model for ASR. We introduce an event-driven SpeechMamba with FATReLU activation, achieving over 60\% activation sparsity with less than 1\% accuracy degradation on LibriSpeech. We also propose a spiking SpeechMamba that attains over 70\% sparsity while using 30\% fewer parameters than comparable SNNs. Finally, we develop a cycle-accurate event-driven simulator enabling flexible algorithm-hardware co-exploration, which helps us identify computational bottlenecks and yields over 10\% additional efficiency improvements.
\end{abstract}


\section{Introduction}
\label{intro}

Deep learning has greatly advanced the performance of automatic speech recognition (ASR) in recent years~\cite{GulatiConformer:Recognition,GaoSPEECH-MAMBA:MODELS}, enabling its widespread adoption in real-world applications. ASR is now increasingly deployed on edge devices~\cite{xu2024conformer}, such as smartphones and smart home devices, where it supports real-time interaction with users. However, the extensive computation required by deep neural networks (DNNs) poses significant challenges for such deployments. The high computational and energy demands of dense matrix multiplication in DNNs limit the applicability of high-performance models on edge devices with strict power budgets, and introduce latency that disrupts natural user interaction~\cite{sze2017efficient}. Therefore, there is a need for computationally efficient solutions for ASR.

Digital neuromorphic computing offers a promising solution to address these efficiency challenges by introducing activation sparsity through spiking neural networks (SNNs)~\cite{eshraghian2023training} or event-based neural networks~\cite{xu2024optimizing}. Unlike conventional DNNs that perform dense matrix multiplications at every layer, neuromorphic approaches convert these operations into sparse computations, where only active neurons contribute to the forward pass. This strategy has been successfully scaled to complex tasks in computer vision~\cite{wang2025context} and language processing~\cite{11173109}. In the speech domain, neuromorphic approaches have demonstrated success in simpler tasks such as keyword spotting~\cite{yang2022deep} and denoising~\cite{sun2024dpsnn}, and recent work has explored large-vocabulary ASR using SNNs for acoustic modeling~\cite{song2025iml}. However, a study that evaluates the realistic hardware benefits of different neuromorphic computing strategies, including spiking and event-driven approaches, for complex ASR remains absent.

Furthermore, existing neuromorphic works in speech processing primarily report efficiency using algorithmic metrics, such as synaptic operations and theoretical energy estimates~\cite{yang2022deep,song2025iml}. These metrics assume ideal sparse computation and do not capture the true costs of execution on real digital neuromorphic hardware, including irregular sparsity that is difficult to exploit efficiently~\cite{liu2022randomize} and memory access patterns with significant data movement overhead~\cite{sze2017efficient}. While existing neuromorphic hardware captures these costs, they impose constraints that limit algorithmic exploration. For example, Loihi~\cite{davies2018loihi} supports only specific neuron models and computation paradigms. This disconnect between abstract metrics and constrained hardware hinders effective algorithm-hardware co-exploration. Algorithmic metrics obscure true hardware bottlenecks, while platform-specific constraints limit the freedom to iterate on network designs.

This paper explores different spike-based and event-driven neuromorphic strategies to improve activation sparsity in the state-of-the-art SpeechMamba model~\cite{GaoSPEECH-MAMBA:MODELS} for complex ASR \footnote{\url{https://github.com/ERNIS-LAB/speech-asr-neuromorphic-mamba}}. Our contributions are as follows:
  \begin{itemize}
  \item We introduce an event-driven SpeechMamba with FATReLU activation function~\cite{Kurtz2020InducingInference}. To maximize activation sparsity while minimizing accuracy degradation, we propose a multi-stage training pipeline. The resulting model achieves an average activation sparsity over 60\% with minimal accuracy loss ($<$1\%) on the LibriSpeech dataset.
  \item We propose a spiking SpeechMamba using binary spikes for computation. With sparsity-aware training, the spiking network achieves over 70\% activation sparsity and delivers competitive performance compared to state-of-the-art SNNs while using 30\% fewer parameters.
  \item We develop a neuromorphic simulator for cycle-accurate event-driven dataflow execution on a RISC-V Ibex core~\cite{LowRISC/ibex:Zero-riscy}. Unlike existing simulators that focus on analytical modeling or coarse event-driven scheduling, our approach provides hardware-aware yet flexible simulation for neuromorphic workloads. Using this simulator, we identify computational bottlenecks in event-driven SpeechMamba and improve its efficiency by over 10\%.
  \end{itemize}

\section{Background and Related Work}
\label{background}

\subsection{Deep Learning Advances for Speech Recognition}

Transformer-based architectures have significantly advanced speech recognition by effectively modeling long-range dependencies in sequential data~\cite{GulatiConformer:Recognition}. However, self-attention mechanisms lack explicit state compression and incur quadratic computational complexity with respect to sequence length~\cite{Bahdanau2014NeuralTranslate}. In contrast, state space models (SSMs)~\cite{GuMamba:Spaces} address these limitations by maintaining compact, input-dependent latent states, enabling more efficient long-context modeling. Recent studies demonstrate that SSM-based speech models can achieve lower word error rates with fewer parameters and reduced computational cost compared to transformer-only architectures~\cite{Lin2025AnModels}.

SpeechMamba~\cite{GaoSPEECH-MAMBA:MODELS} introduced a speech recognition architecture that integrates Mamba blocks with self-attention. The model adopts an encoder-decoder architecture, which consists of multiple repeated units. Within each unit, two Mamba blocks are interleaved with self-attention layers. While the Mamba blocks capture long-range temporal dependencies, the self-attention layers model lower-level temporal representations. Owing to its compact architecture and strong speech recognition performance, we select SpeechMamba as the foundational architecture supporting our proposed approaches.

\subsection{Attention-based Spiking Neural Networks}

Spikformer~\cite{ZhouSPIKFORMER:TRANSFORMER} and SpikMamba~\cite{ChenSpikMamba:Format} are recent spiking neural architectures that combine the efficiency of SNNs with attention-based sequence modeling. Spikformer introduces Spiking Self Attention (SSA), where Query, Key, and Value are represented as spikes and softmax is replaced with sparse, multiplication-free operations. SpikMamba incorporates spiking Mamba blocks and a spiking window-based linear attention mechanism to capture both global and local temporal dependencies in event-based data. In our work, we adopt SSA for the spiking attention module and leverage spiking Mamba blocks as the foundation of our spiking SpeechMamba.

\subsection{Event-driven Digital Neuromorphic Processing}

Event-driven digital neuromorphic processing exploits activation sparsity by computing and communicating only non-zero activations. When activations fall below a threshold, downstream computations and memory accesses are skipped entirely, where cost scales with active neurons rather than total network size~\cite{xu2024optimizing}. This paradigm originates from spiking neural networks, where neurons emit binary spikes only when their membrane potential exceeds a firing threshold, naturally inducing high sparsity. Recent work has generalized beyond binary spikes to graded activations, retaining the computational benefits of event-driven processing while avoiding quantization error inherent in binary spikes~\cite{xu2024optimizing,wang2025context,11173109}. From a hardware perspective, early neuromorphic architectures stored per-neuron states on-chip, incurring significant memory overhead~\cite{davies2018loihi}. Current designs address this through hybrid configurations combining stateful spiking layers with non-stateful event-based layers~\cite{Yousefzadeh2022SENeCA:Architecture}. These advances make high-sparsity networks increasingly suitable for memory-constrained neuromorphic deployment.

\subsection{Simulators for Event-driven Neuromorphic Computing}

The ideal approach for evaluating neuromorphic algorithms is to deploy networks directly on existing digital neuromorphic processors. While effective for small networks composed of standardized operators supported by the hardware, this approach becomes problematic when algorithms employ novel architectures or operations not yet supported by existing chip designs. In such cases, hardware-aware simulators are essential to estimate performance and enable algorithm-hardware co-exploration. However, current simulation approaches have limitations. First, analytical simulators, such as Timeloop~\cite{parashar2019timeloop}, offering analytical insight but omit essential hardware metrics such as latency, energy, and cycle counts. Second, hardware-accurate simulators deliver accurate system-level data but remain inflexible, hindering rapid architectural exploration. For instance, custom accelerator architectures~\cite{9937422}, compute-in-memory frameworks~\cite{chauvaux2025event}, and coarse event-driven scheduling schemes~\cite{gimenes2025ample}, where each event typically represents a complete real-time input (e.g., an entire image frame). PyCARL~\cite{balaji2020pycarl} integrated cycle-accurate models of specific neuromorphic chips (e.g., Loihi) with the CARLsim simulator. However, it remains constrained to predefined chip architectures and does not provide a flexible framework for novel architectures. To the best of our knowledge, no existing approach supports a flexible, operation-level, event-driven simulation framework specifically tailored to neuromorphic workloads that process atomic input events, such as individual binary/graded spikes or pixels from event-driven sensors.

\section{Method}

\subsection{Event-driven SpeechMamba}

\begin{figure*}[h]
  \includegraphics[width=\textwidth]{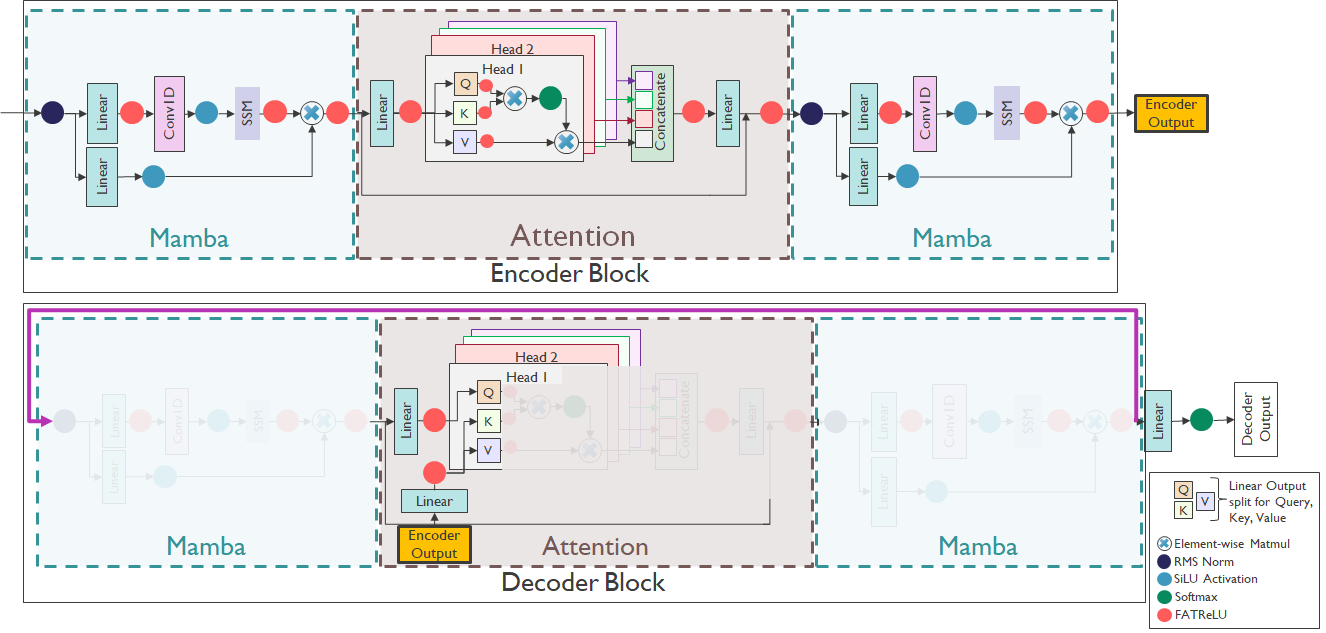}
  \vspace*{-4mm}
  \caption{Encoder and decoder blocks of the E-SpeechMamba architecture. Red dots indicate inserted FATReLU activation points. Dimmed modules in the decoder block are identical to their encoder counterparts.}
  \label{fig:reluencoderblock}
\end{figure*}

We propose event-driven SpeechMamba (E-SpeechMamba) with FATReLU-induced activation sparsity at key points in the SpeechMamba architecture~\cite{GaoSPEECH-MAMBA:MODELS}, which reduces the computational load on the downstream operations. To effectively introduce activation sparsity for event-driven computation with minimal accuracy degradation, we propose a three-stage sparsification pipeline, including ReLU pre-training, FATReLU threshold initialization, and FATReLU threshold finetuning.

Firstly, we introduce activation sparsity by inserting ReLU activations at key bottleneck points of SpeechMamba during pre-training. In addition, a sparsity loss function is employed to train E-SpeechMamba. With these configurations, the computational load of SpeechMamba is significantly reduced while maintaining model performance. As illustrated in Fig.~\ref{fig:reluencoderblock}, a ReLU activation is added after each linear layer, convolutional layer, and SSM block. Within the SSM blocks, the insertion points are selected to preserve the learning dynamics of the SSM matrices $A$, $B$, $C$, and $D$, which typically parameterize orthogonal basis functions, such as Legendre polynomials~\cite{VoelkerLegendreNetworks}, with both positive and negative values.

Secondly, we replaced all ReLU activations with FATReLU~\cite{Kurtz2020InducingInference} in E-SpeechMamba to further improve the activation sparsity. FATReLU is a parameterized variant of ReLU that enforces sparsity by suppressing activations below a learnable threshold \(T\). Formally, for an activation \(x\), FATReLU is defined as
\[
\text{FATReLU}(x) =
\begin{cases}
x & \text{if } x \ge T, \\
0 & \text{otherwise},
\end{cases}
\]
where the threshold \(T > 0\) is manually initialized.

Following the method of~\cite{Kurtz2020InducingInference}, the threshold initialization estimates an initial threshold for each FATReLU layer using a data-driven threshold sweeping strategy. Specifically, inference is first performed using the E-SpeechMamba model from the first step, on a representative batch of training data to obtain a reference loss, referred to as the \emph{base\_loss}. Activation statistics are then collected at each FATReLU insertion point. The initial threshold \(T\) is set to the mean of the lowest 10\% (i.e., the first decile) of the recorded activation values. E-SpeechMamba is subsequently re-evaluated with these thresholds applied to compute an \emph{updated\_loss}.

The threshold is then progressively increased by sweeping over higher activation deciles. At each step, the updated loss is compared against the base loss, and the threshold increase is accepted as long as the normalized loss ratio
\[
\frac{\text{updated\_loss}}{\text{base\_loss}}
\]
remains below a predefined tolerance \(K\). This threshold sweeping process enables aggressive activation sparsification while explicitly constraining the performance degradation.

Lastly, after initializing the thresholds \(T\), we finetune them during training by incorporating an additional sparsity-inducing regularization term into the base loss. Specifically, the sparsity loss is defined as
\begin{equation}
L_{\mathrm{spar}} = \sum_{i} \left( \text{FATReLU}(\mathbf{x}_{i}) + \left( \frac{1}{T_{i}} \right)^{2} \right),
\label{eq:lossl11}
\end{equation}
where \(i\) indexes all inserted FATReLU activation layers. This sparsity loss jointly penalizes large activation values and small threshold values, thereby encouraging more activations to be forced to zero and promoting event-driven sparsity~\cite{XuEvent-basedO}.

\subsection{Spiking SpeechMamba}

Additionally, we propose the Spiking SpeechMamba (S-SpeechMamba) with leaky integrate-and-fire (LIF) spiking neuron layers. The S-SpeechMamba substitutes the event-driven Mamba modules of E-SpeechMamba, which use FATReLU activations, with the spiking Mamba architecture introduced in SpikMamba~\cite{ChenSpikMamba:Format}. Furthermore, the S-SpeechMamba incorporates the attention block design from Spikformer~\cite{ZhouSPIKFORMER:TRANSFORMER}, where the input features are linearly projected into query, key, and value representations, followed by normalization and LIF layers to generate binary spike trains. Attention is then computed using scaled dot-product similarity between spiking queries and keys, without softmax normalization, and applied to the spiking values to aggregate contextual information. This technique preserves the global dependency modeling of Transformers while improving computational and energy efficiency.

During S-SpeechMamba training, we introduce a firing-rate regularization term to promote sparsity while ensuring effective SNN training. The regularization consists of two complementary components, \(L_{\text{quiet}}\) and \(L_{\text{burst}}\). While \(L_{\text{quiet}}\) penalizes layers with insufficient firing activity to prevent dead neurons during training, \(L_{\text{burst}}\) penalizes excessive firing to increase activation sparsity:
\begin{equation}
\begin{aligned}
L_{\text{quiet}} &= \frac{1}{L} \sum_{l=1}^{L} \max \left( 0, r_{\min} - \bar{r}^{(l)} \right), \\
L_{\text{burst}} &= \frac{1}{L} \sum_{l=1}^{L} \max \left( 0, \bar{r}^{(l)} - r_{\max} \right),
\end{aligned}
\end{equation}
where \(\bar{r}^{(l)}\) denotes the average firing rate of neurons in spiking layer \(l\), and \(r_{\min}\) and \(r_{\max}\) represent the minimum and maximum target firing rates, respectively.

The overall firing-rate regularization loss is defined as a summation of the two components.

\subsection{Event-driven Neuromorphic Simulator}

We develop a neuromorphic simulator enabling operation-level, event-driven dataflow execution on a RISC-V Ibex core~\cite{LowRISC/ibex:Zero-riscy}, emulating generic digital neuromorphic processors like SENECA~\cite{Yousefzadeh2022SENeCA:Architecture} and SpiNNaker2~\cite{Mayr2019SpiNNakerLearning}. The simulator models fine-grained processing of atomic inputs through partial-sum computation within network layers, capturing the benefits of unstructured activation sparsity. It provides a hardware-aware yet programmable environment that supports rapid, faithful exploration of neuromorphic architectures while accurately reflecting key system-level metrics.

We adapt both E-SpeechMamba and S-SpeechMamba for the event-driven dataflow simulator. Each layer generates partial outputs and forwards them immediately to subsequent layers as soon as the corresponding atomic events become available. For SNNs, the simulator provides for additional computational efficiency achieved by representing spikes and non-spikes as integer values 1 and 0. This representation allows the simulator to bypass energy-intensive multiplications involving binary activations and real-valued weights in multiply-accumulate (MAC) operations. Specifically, multiplications with binary zeros are skipped, while multiplications with binary ones directly propagate the corresponding real-valued weights, thereby reducing computational overhead during event-driven execution.

\begin{figure}[h]
  \centering
  \includegraphics[width=0.46\textwidth]{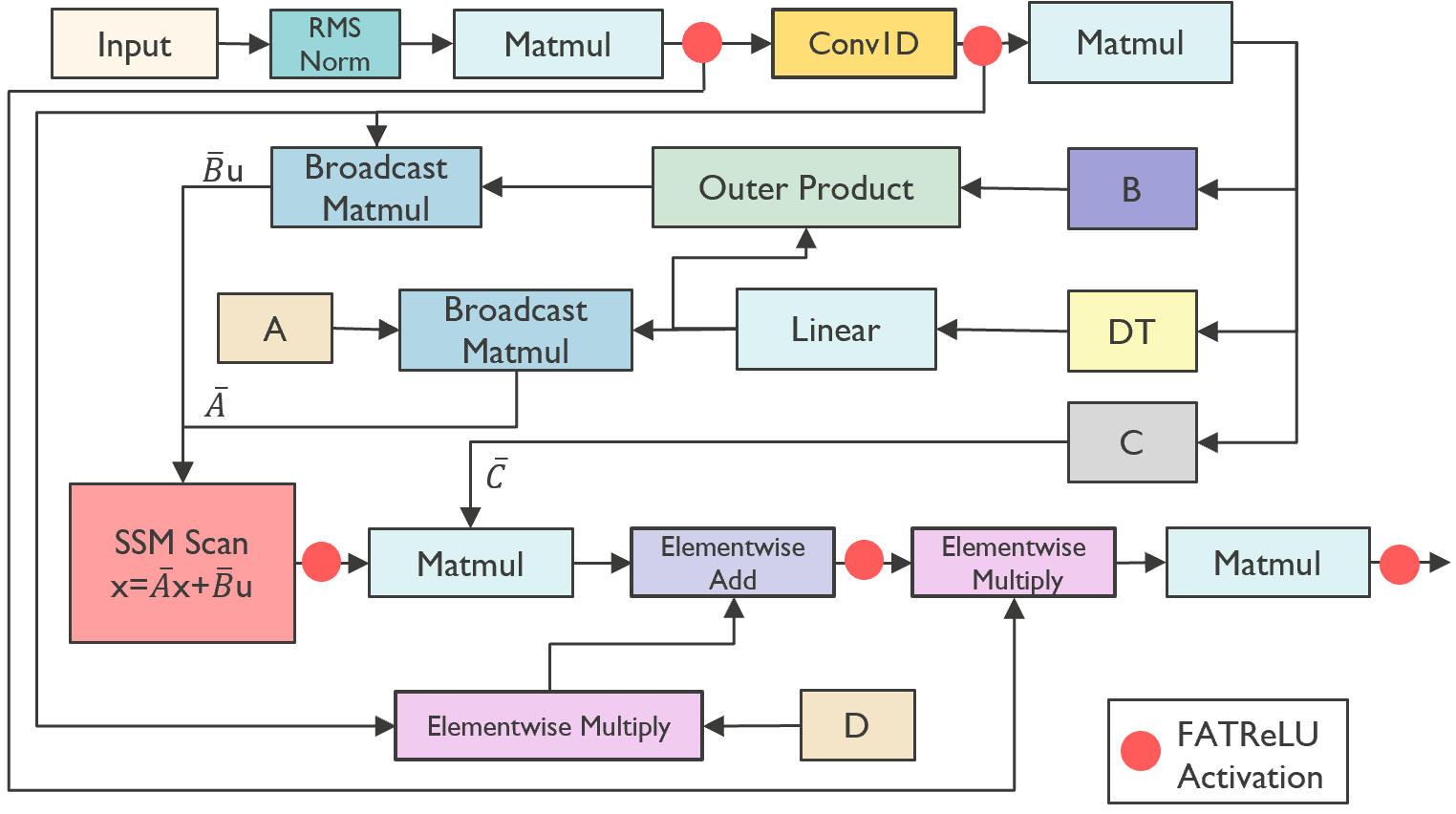}
  \caption{Block diagram of Mamba block in E-SpeechMamba (Optimized) with added FATReLU activation points.}
  \label{fig:improvedann}
\end{figure}

These event-driven networks are subsequently lowered onto a simple Ibex-based system and compiled using the lowRISC GCC-based toolchain\cite{LowRISC/ibex:Zero-riscy}. The resulting binaries are executed on the Ibex simulator, which models an RV32IMC Ibex core with a unified instruction and data memory. The simulator is implemented using Verilator and provides detailed performance counters.

\subsection{Simulator-driven E-SpeechMamba Optimization}

After the simulation, performance analysis is performed to help us identify efficiency bottlenecks. By jointly considering these bottlenecks and the structural properties of the SSMs, we introduce additional FATReLU activation points in the Mamba block to further optimize the efficiency of E-SpeechMamba, shown as the E-SpeechMamba (Optimized) in Fig. \ref{fig:improvedann}.

\section{Experiment and Results}

\begin{table*}[t]
\centering
\caption{Comparison with state-of-the-art methods on LibriSpeech. Lower values are better for all metrics except sparsity.}
\begin{tabular}{lcccccccc}
\hline
Model & \multicolumn{4}{c}{WER (\%)} & \# Params (M) & Sparsity (\%) \\ \cline{2-5}
      & dev-clean & dev-other & test-clean & test-other &  &  \\ 
\hline
Whisper-Large-V2\cite{RadfordRobustSupervision}
      & -- & -- & 2.7 & 5.2 & 1550 & -- \\
Pruned Conformer\cite{JiangAccurateRecognition}
      & -- & -- & 3.27 & 6.89 & 71.5 & 50 \\
SpeechMamba\cite{GaoSPEECH-MAMBA:MODELS}
      & 2.16 & 5.13 & 2.32 & 5.23 & 67.6 & -- \\
\hline
Spike-driven Transformer\cite{yao2023spike}
      & 8.7 & 20.7 & 8.9 & 22.3 & 99.4 & -- \\
IML-Spikeformer\cite{song2025iml}
      & 3.1 & 8.3 & 3.4 & 7.9 & 99.4 & -- \\
\hline
\shortstack{Base Model (SpeechMamba)\\ with naive sparsification}
      & 2.30 & 5.51 & 2.47 & 5.86 & 67.6 & 20 \\
\textbf{E-SpeechMamba}
      & 2.90 & 7.40 & 3.20 & 7.80 & 67.6 & 62 \\
\textbf{S-SpeechMamba}
      & 4.27 & 9.34 & 4.71 & 9.98 & 67.8 & 72 \\
\textbf{E-SpeechMamba (Optimized)}
      & 3.10 & 7.80 & 3.60 & 8.30 & 67.6 & 64 \\ \hline
\end{tabular}
\label{tab:main-result}
\end{table*}

Our experiments are designed to achieve three key objectives. First, we demonstrate our sparse neuromorphic SpeechMamba models by benchmarking their performance against state-of-the-art approaches and comparing the performance and sparsity characteristics of different neuromorphic strategies. Second, we analyze model performance on our neuromorphic simulator and investigate the gap between algorithmic sparsity and hardware performance for both event-driven and spiking models. Third, we conduct a module-level analysis supported by simulation results to explain how we further optimize our event-driven SpeechMamba model to achieve greater hardware efficiency gains. All experiments are conducted on the LibriSpeech dataset \cite{PanayotovLIBRISPEECH:BOOKS}, a large-scale corpus of approximately 1,000 hours of read English speech sampled at 16 kHz. We used all training splits (approximately 980 hours) of the dataset to train our models.

\subsection{Experiment Setup}

Pre-training of E-SpeechMamba, E-SpeechMamba (Optimized), and training of S-SpeechMamba were conducted using the SpeechBrain toolkit\cite{Ravanelli2021SpeechBrain:Toolkit} on a single NVIDIA A100 GPU. For reproducibility, the random seed was fixed to 74443 across all experiments, and full-precision (FP32) training was employed throughout. All models were trained for 100 epochs using the Adam optimizer with parameters $\beta_1 = 0.9$, $\beta_2 = 0.98$, and $\epsilon = 10^{-9}$. The initial learning rate was set to $1\times10^{-3}$ and scheduled using the Noam learning rate scheduler with 25,000 warm-up steps. Gradient accumulation with a factor of 4 was applied to effectively increase the batch size, and gradient norms were clipped to a maximum value of 5.0. Dynamic batching was enabled to accommodate variable-length utterances, with the maximum batch length capped at 1024 frames during training. Threshold initialization for E-SpeechMamba and E-SpeechMamba (Optimized) was performed on an Intel Xeon–based CPU, after which the models were finetuned on the training set for an additional 20 epochs. Furthermore, event-driven simulations were executed on an Intel Core i7-9750H CPU hosting the Ibex simulator.

\subsection{Benchmarking sparse neuromorphic SpeechMamba models against the state-of-the-arts}

Table \ref{tab:main-result} compares our neuromorphic SpeechMamba models against state-of-the-arts on LibriSpeech evaluation splits, each approximately 5 hours in duration. The "clean" subsets represent clearer acoustic conditions, while "other" subsets contain more challenging samples with greater speaker variability. We report Word Error Rate (WER), which indicates the percentage of incorrectly recognized words. Our E-SpeechMamba achieves 62\% sparsity with less than 1\% WER increase on test-clean compared to the dense baseline \cite{GaoSPEECH-MAMBA:MODELS}. The S-SpeechMamba attains the highest averaged sparsity, benefiting from efficient binary-spike computation despite higher WER. The S-SpeechMamba also delivers competitive performance compared with state-of-the-art SNN solutions while using 30\% less parameters. The optimized E-SpeechMamba achieves similar average sparsity to the original version. However, this sparsity is computed over more activation points, resulting in greater overall computation reduction.

\begin{figure}[h]
  \centering
  \includegraphics[width=0.45\textwidth]{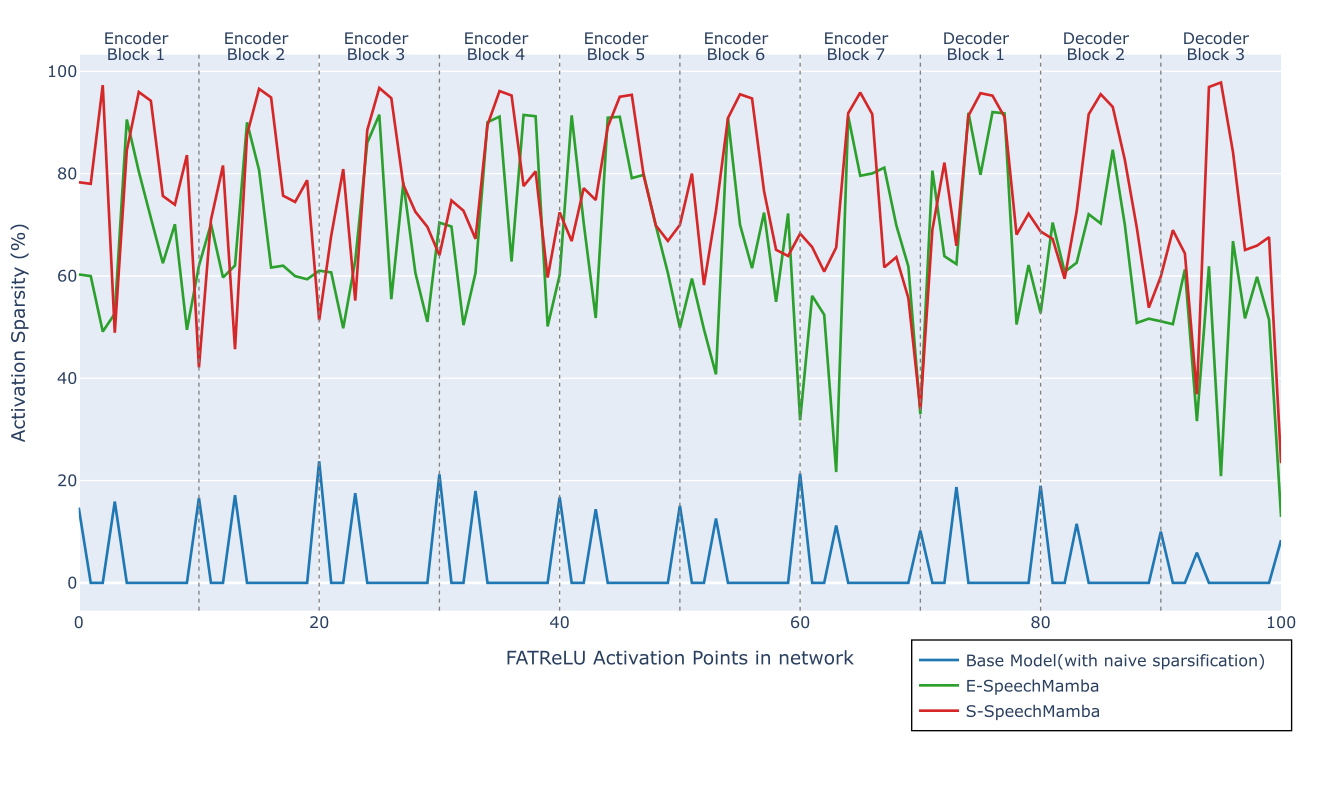}
  \caption{Activation sparsity within each encoder and decoder block of E-SpeechMamba and S-SpeechMamba models at each activation sparsity point introduced in Fig. \ref{fig:reluencoderblock}.}
  \label{fig:result-sparsity}
\end{figure}

Fig. \ref{fig:result-sparsity} examines the internal activation sparsity of our models compared to the base model, which applies ReLU only at the end of each module, using the first 10\% of the test-clean data. The results reveal substantial variation across sparsification points, ranging from over 90\% at certain locations to below 40\% at others. The sparsity distribution suggests that hardware deployment could benefit from tailored strategies for different activation points within the network.

\subsection{Event-driven Simulation}

We quantitatively evaluate the benefits of our methods by simulating the models on our event-driven simulator. We measure the following hardware performance metrics:
\begin{itemize}\setlength{\itemsep}{0pt}\setlength{\topsep}{0pt}
\item \textit{CPU cycles}: Clock cycles consumed by the event-driven simulator to execute the model.
\item \textit{CPU instructions (inst.)}: Total instructions executed by the simulator.
\item \textit{Memory access (acc.)}: Number of memory loads and stores in the simulation.
\item \textit{Latency}: Average processing time per input sample.
\end{itemize}
Using the naively sparsified base model from Table \ref{tab:main-result} as our baseline, Table \ref{tab:hardwarespecs} reports the percentage improvement of our proposed methods across these metrics.

\begin{table}[h]
\centering
\caption{Percentage improvement in simulated hardware efficiency metrics relative to the base model}
\begin{tabular}{lcccc}
\hline
Model               & \shortstack{CPU\\Cycles} & \shortstack{CPU\\Inst.} & \shortstack{Memory\\Acc.} &Latency \\ \hline
E-SpeechMamba          & 32.32      & 14.30            & 17.57 & 29.78 \\
S-SpeechMamba                 & 19.58      & 14.0                & 7.63 & 17.9 \\
E-SpeechMamba (Optimized) & 46.13      & 26.9             & 28.50 & 37.5 \\ \hline
\end{tabular}
\label{tab:hardwarespecs}
\end{table}

The simulation results provide deeper insight into real hardware performance beyond average sparsity metrics. Table \ref{tab:hardwarespecs} shows that the CPU cycle improvement of E-SpeechMamba does not directly correspond to the reported sparsity levels. This discrepancy arises from two factors. First, each FATReLU point exhibits different sparsity levels and operates on varying dimensions, resulting in uneven computational benefits. Several high-dimensional points with relatively lower activation sparsity limit the overall CPU cycle reduction. Second, certain critical operations within the Mamba block cannot be sparsified. For instance, sparsifying the input to the SSM Scan module causes significant accuracy degradation. These remaining dense operations, which are not reflected in the algorithmic sparsity analysis, still contribute substantially to the overall computation.

Surprisingly, Table \ref{tab:hardwarespecs} shows that S-SpeechMamba achieves lower CPU cycle reduction than E-SpeechMamba, despite having higher activation sparsity. This reveals a hidden cost of spiking neurons often overlooked in neuromorphic research: the overhead of maintaining membrane state. Unlike stateless activations, LIF neurons require additional memory loads and stores to preserve membrane potential across time steps. This is evidenced by the over 10\% lower memory access improvement of S-SpeechMamba compared to E-SpeechMamba in Table \ref{tab:hardwarespecs}. Therefore, event-driven approaches may be preferable when memory bandwidth is the primary constraint, despite their lower activation sparsity.

\subsection{Simulation Analysis of Submodules with Mamba Block}

\begin{figure}[h]
    \centering

    \begin{subfigure}{0.45\textwidth}
        \centering
        \includegraphics[width=\linewidth]{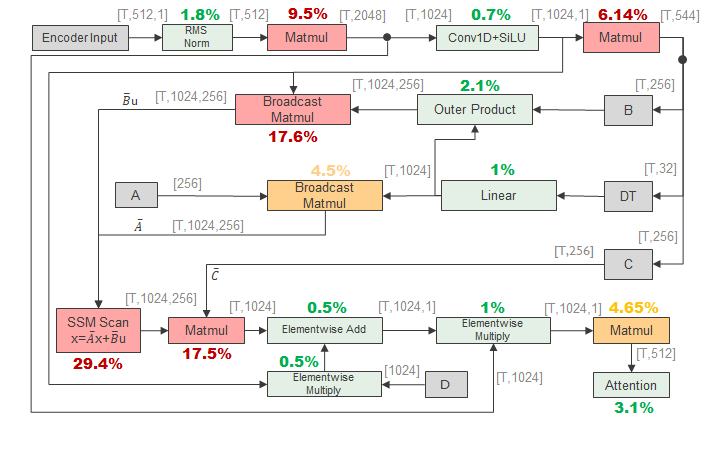}
        \vspace*{-7mm}
        \caption{Base model with naive sparsification.}
    \end{subfigure}
    
    \begin{subfigure}{0.45\textwidth}
        \centering
        \includegraphics[width=\linewidth]{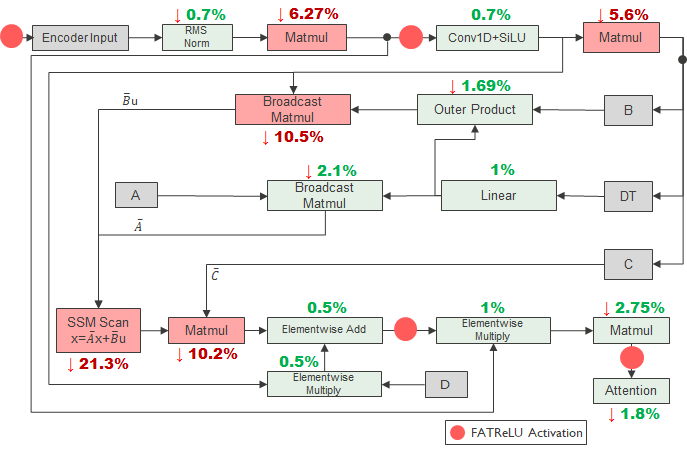}
        \vspace*{-5mm}
        \caption{E-SpeechMamba (Normalized to base model.)}
    \end{subfigure}

    \begin{subfigure}{0.45\textwidth}
        \centering
        \includegraphics[width=\linewidth]{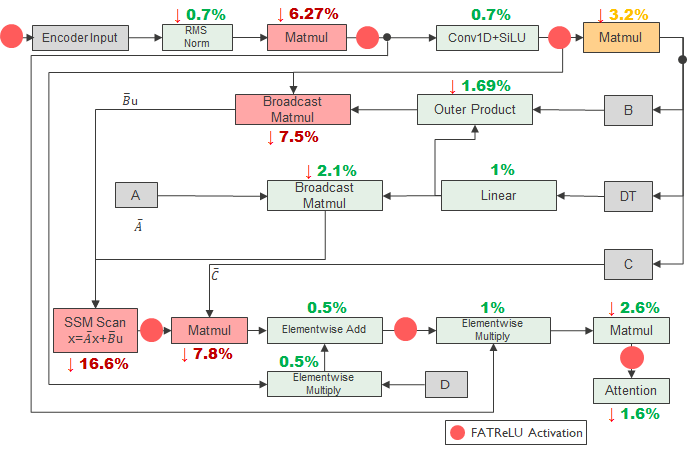}
        \vspace*{-5mm}
        \caption{E-SpeechMamba (Optimized) (Normalized to base model.)}
    \end{subfigure}

    \caption{Simulated CPU cycles distributions across different submodules within the encoder Mamba block.}
    \label{fig:perfcompare}
\end{figure}

Our simulator enables performance measurement at the submodule level within each block. Fig. \ref{fig:perfcompare} shows the CPU cycle distribution across detailed submodules within the encoder Mamba block. The percentage for each submodule is computed by dividing its executed cycles by the total cycles of the base model for the Mamba block. This normalization allows direct comparison of CPU cycle reduction across submodules for different model variants.

From Fig. \ref{fig:perfcompare}(b), we identify remaining computational hotspots within the Mamba module, including the SSM Scan submodule and several matrix multiplication operations, which continue to dominate execution time. These findings motivated further optimization of E-SpeechMamba by introducing additional FATReLU sparsification points. As shown in Fig. \ref{fig:perfcompare}(c), the optimized model substantially reduces CPU cycles at these identified hotspots, yielding the significant hardware efficiency improvements reported in Table \ref{tab:hardwarespecs}. This iterative approach of profiling, hotspot identification, and targeted optimization provides a generalizable framework for simulator-based algorithm-hardware co-exploration.

\section{Discussion and Conclusion}

In this paper, we present spiking and event-driven neuromorphic approaches to improve activation sparsity in the SpeechMamba model for complex ASR tasks. We introduce an event-driven SpeechMamba employing the FATReLU activation function, supported by a multi-stage training pipeline to maximize sparsity while preserving accuracy. We further propose a spiking SpeechMamba using binary spikes with sparsity-aware training. To enable algorithm-hardware co-exploration, we developed a neuromorphic simulator supporting cycle-accurate, event-driven dataflow execution on a RISC-V Ibex core. Our simulator provides flexible, operation-level simulation for the workloads, enabling our identification of computational bottlenecks and guiding targeted optimizations.

Our simulation analysis reveals a mismatch between algorithmic sparsity and hardware performance, where inefficiencies in memory access patterns and uneven sparsity can substantially erode expected gains in latency and energy. This underscores the importance of hardware-aware analysis, not only to validate proposed methods under realistic conditions, but also to identify bottlenecks guiding further optimization. Neuromorphic algorithm and hardware research have historically progressed independently, with novel algorithms often advancing in directions that existing hardware cannot efficiently support, obscuring whether algorithmic improvements yield tangible hardware benefits. Simulation tools bridging this gap are essential for accelerating progress in the field.

\section*{Acknowledgment}
This publication is part of the project Brain-inspired MatMul-free Deep Learning for Sustainable AI on Neuromorphic Processor with file number NGF.1609.243.044 of the research programme AiNed XS Europe which is (partly) financed by the Dutch Research Council (NWO) under the grant https://doi.org/10.61686/MYMVX53467. This work was also supported by the AI competence center ScaDS.AI Dresden/Leipzig (01IS18026A-D) and by the European Horizon MSCA Doctoral Network REACT "Self-AwaRe NEuromorphic ArChiTectures: Security, Reliability and Energy-Efficiency" (101226463).

\bibliography{reference}
\bibliographystyle{IEEEtran}

\end{document}